\documentclass[runningheads]{llncs}
\usepackage{graphicx}
\usepackage{comment}
\usepackage{amsmath,amssymb} 
\usepackage{color}
\usepackage{booktabs}
\usepackage{array}
\usepackage{diagbox}
\usepackage{caption}
\usepackage{authblk}
\usepackage{algorithm}
\usepackage{algorithmic}
\usepackage{subcaption}
\usepackage{caption}
\usepackage{bm}
\usepackage[breaklinks=true,bookmarks=false]{hyperref}
\newcolumntype{C}[1]{>{\centering\arraybackslash}p{#1}}
\begin{document}

\pagestyle{headings}
\mainmatter
\def\ECCVSubNumber{6}  

\title{Addressing Neural Network Robustness with Mixup and Targeted Labeling Adversarial Training} 

\titlerunning{Addressing Robustness with M-TLAT}
%
\author{Alfred LAUGROS\inst{1,2} \and
Alice CAPLIER\inst{2} \and
Matthieu OSPICI\inst{1}}
\authorrunning{A. Laugros et al.}
%
\institute {Atos, France
\email{\{alfred.laugros,matthieu.ospici\}@atos.net} \and
Univ. Grenoble Alpes, France \email{alice.caplier@grenoble-inp.fr}}

\maketitle

\begin{abstract}
Despite their performance, Artificial Neural Networks are not reliable enough for most of industrial applications. They are sensitive to noises, rotations, blurs and adversarial examples. There is a need to build defenses that protect against a wide range of perturbations, covering the most traditional common corruptions and adversarial examples. We propose a new data augmentation strategy called M-TLAT and designed to address robustness in a broad sense. Our approach combines the Mixup augmentation and a new adversarial training algorithm called Targeted Labeling Adversarial Training (TLAT). The idea of TLAT is to interpolate the target labels of adversarial examples with the ground-truth labels.  We show that M-TLAT can increase the robustness of image classifiers towards nineteen common corruptions and five adversarial attacks, without reducing the accuracy on clean samples.

\keywords{Neural Network, Robustness, Common Corruptions, Adversarial Training, Mixup}
\end{abstract}

\section{Introduction}

Artificial neural networks have been proven to be very efficient in various image processing tasks \cite{imagenet},\cite{rcnn},\cite{face_rec_ref}. Unfortunately, in real-world computer vision applications, a lot of common corruptions may be encountered such as blurs, colorimetry variations, or noises, etc. Such corruptions can dramatically decrease neural network efficiency \cite{rot_transl},\cite{noise_vul},\cite{face_rec_noise},\cite{blur}. Besides, deep neural networks can be easily attacked with adversarial examples \cite{ref_adv}. These attacks can reduce the performance of most of the state-of-the-art neural networks to zero \cite{adv_phy},\cite{practical_adv}.

Some techniques have been proposed to make neural networks more robust. When a specific perturbation is considered, we can build a defense to protect a neural network against it, whether it is a common corruption \cite{add_noise_study},\cite{fine_tune_noise},\cite{blur} or an adversarial attack \cite{pgd_aug},\cite{squeez},\cite{defense_gan}. However, increasing robustness towards a specific perturbation generally does not help with another kind of perturbation. For instance, geometric transformation robustness is orthogonal with worst-case additive noise \cite{spatial_rob}. Fine tuning on blur does not increase the robustness to Gaussian noise \cite{mixt_expert}. Even worse, making a model robust to one specific perturbation can make it more sensitive to another one. For instance, a data augmentation with corruptions located in the high frequency domain tends to decrease the robustness to corruptions located in the low frequency domain \cite{fourier}. Besides, increasing adversarial robustness often implies a diminution of the \textit{clean accuracy} (the accuracy of a model on not-corrupted samples) \cite{adv_scale},\cite{odds_with}. Therefore, a company or a public organism may feel reluctant to use neural networks in their projects since it is hard to make them robust to a large diversity of corruptions, and it is difficult to predict how a neural network will react to unexpected corruptions. There is a need to build new defenses that address robustness in a broad sense, covering the most encountered common corruptions and adversarial examples.\\

We propose to address this issue with a new data augmentation approach called M-TLAT. M-TLAT is a combination of Mixup \cite{mixup} and a new kind of adversarial training called Targeted Labeling Adversarial Training (TLAT). The idea of this adversarial training is to label target adversarial examples with soft labels that contain information about the used target. We show that M-TLAT can increase the robustness of image classifiers to nineteen common corruptions and five adversarial attacks, without reducing the accuracy on clean samples. This algorithm is easy to implement and to integrate to an existing training process. It intends to make the neural networks used in real-world applications more reliable.

\section{Related Works}

\subsection{Protecting Neural Networks against Common Corruptions}

Neural Networks are known to be sensitive to a lot of perturbations such as noises \cite{add_noise_study}, rotations \cite{rot_transl}, blurs \cite{blur} or colorimetry variations \cite{face_rec_noise}, etc. We call these perturbations common corruptions. They are often encountered in industrial applications, but generally absent from academic datasets. For instance the faces in the dataset celeba are always well illuminated with a consistent eyes positioning \cite{celeba}, yet those conditions are not always guaranteed in the industrial applications. Because of common corruptions, the performance of a neural network can be surprisingly low \cite{compair_human}.

There are a few methods that succeded in increasing the robustness to several common corruptions simoultaneously. Among them, the robust pre-training algorithm proposed by Liu et al. can make classifiers more robust to noises, occlusions and blurs \cite{liu_robust}. In \cite{stylized_imagenet}, it is proposed to change the style of the training images using style transfer \cite{style_transfer}. Neural networks trained this way are obliged to focus more on shapes than textures. The Augmix algorithm proposes to interpolate a clean image with perturbed versions of this image \cite{augmix}. The obtained images are used to augment the training set of a neural network. These methods are useful to make neural networks more robust to common corruptions but they do not address the case of adversarial examples.

\subsection{Protecting Neural Networks against Adversarial Examples}
Adversarial Examples are another threat that can make neural networks give unexpected answers \cite{ref_adv}. Unlike common corruptions they are artificial distortions. They are crafted by humans so as to especially fool neural networks. Adversarial examples can completely fool even the state-of-the-art models \cite{adv_scale}. Those perturbations are even more dangerous because humans can hardly see if an image has been adversarially corrupted or not \cite{adv_phy},\cite{practical_adv}. In other words, a system can be attacked without anyone noticing it. They are two kinds of adversarial examples called white-box and black-box attacks.

\textbf{White-box attacks.}
The attacker has access to the whole target network: its parameters and its architecture. White-box attacks are tailored so as to especially fool a specific network. White-box adversarial examples are very harmful, defending a model against it is a tough task \cite{odds_with},\cite{adv_more_data},\cite{obfu_gradient}.

\textbf{Black-Box attacks.}
An adversarial example crafted with a limited access to the targeted network is called a black-box attack. When only the training set of a neural network is known, it is still possible to make a transfer attack. Considering a dataset and two neural networks trained with it, it has been shown that an adversarial example crafted using one of the models, can harm the other one \cite{ref_adv},\cite{harnessing}. This phenomenon occurs even when the two models have distinct architectures and parameters.\\

A lot of methods have been proposed to protect against adversarial examples. Adversarial training uses adversarial examples to augment the training set of a neural network \cite{pgd_aug},\cite{ensemble_training}. Defense-Gan \cite{defense_gan} and feature squeezing \cite{squeez} are used to remove adversarial patterns from images. Stochastic Activation Pruning \cite{stochastic_prun} and defensive dropout \cite{def_drop} make the internal operations of neural networks more difficult to access in order to make these networks more difficult to attack. These methods can significantly increase the robustness of neural networks to adversarial examples but they do not provide any protection towards common corruptions.

\subsection{Addressing Robustness in a Broad Sense}

The methods mentioned above increase either the adversarial robustness or the robustness to common corruptions. Unfortunately, increasing the robustness to common corruptions generally does not imply increasing the robustness to adversarial examples and conversely \cite{laugros19}. The experiments carried out in \cite{fourier} show that data augmentation with traditional adversarial examples makes models less robust to low frequency corruptions. Robustness to translations and rotations is independent from robustness to the $L_{p}$ bounded adversarial examples \cite{spatial_rob}. 

A natural approach to address robustness in a broad sense is to combine defenses that address common corruptions with defenses that address adversarial examples. Unfortunately, it is possible that two defenses are not compatible and do not combine well \cite{multi_adv_rob},\cite{fourier},\cite{unforeseen}.

A few standalone methods have been recently proposed to address adversarial robustness and robustness to common corruptions at the same time. In \cite{rob_resnext},\cite{self_training}, a large set of unlabeled data are leveraged to get a significant increase in common corruption robustness and a limited increase in adversarial robustness. However, using these methods has a prohibitive computational cost. Adversarial Noise Propagation adds adversarial noise into the hidden layers of neural networks during the training phase to address both robustnesses \cite{ANP}. Adversarial Logit Pairing (ALP) encourages trained models to output similar logits for adversarial examples and their clean counterparts \cite{adv_logit_pairing}. In addition to the adversarial robustness provided, it has been reported that ALP increases the robustness to some common corruptions \cite{imagenet_c}. The drawback of these methods is that they reduce the clean accuracy of trained models.

To be useful in real-world applications, we want our method to preserve the clean accuracy of the trained models, to increase the robustness to both adversarial examples and common corruptions, and to be easy to integrate into an existing training framework.

\section{Combining Mixup with Targeted Labeling Adversarial Training: M-TLAT}

\begin{figure}
  \centering
  \includegraphics[width=\textwidth]{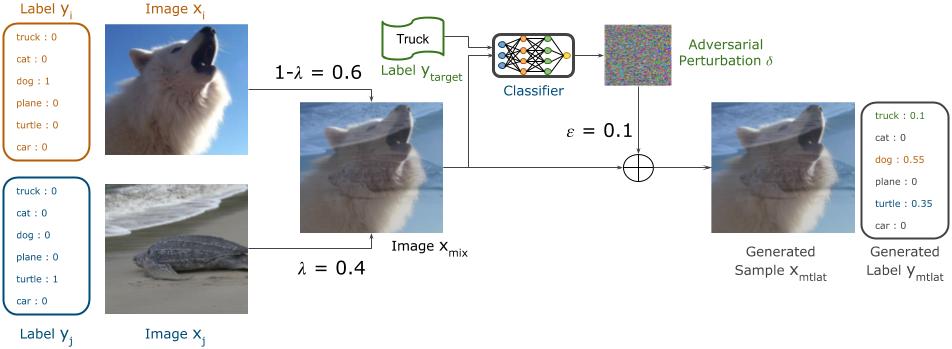}
  \label{fig:benchmark}
  \caption{Visualising of the generation of a new training couple using M-TLAT \label{fig}}
\end{figure}

Our approach called M-TLAT is a combination of two data augmentation algorithms, which are Mixup \cite{mixup} and a new adversarial training strategy called Targeted Labeling Adversarial Training (TLAT). Basically, Mixup aims to increase the common corruption robustness while TLAT aims to increase the adversarial robustness. We go more into details in Section \ref{ablation_study} to understand the contribution of each component. In practice, we observe that those augmentations combine well to address robustness in a broad sense.

\subsection{Mixup}

Let us consider the couples $(x_i,y_i)$ and $(x_j,y_j)$, where $x_i$ and $x_j$  are images of the training set and $y_i$ and $y_j$ are their associated one-hot encoding labels. Mixup \cite{mixup} is a data augmentation algorithm that interpolates linearly samples and labels (see Figure \ref{fig}):

\begin{equation}\label{mixup}
\begin{split}
x_{mix} & = \lambda*x_i + (1-\lambda)*x_j\\
y_{mix} & = \lambda*y_i + (1-\lambda)*y_j
\end{split}
\end{equation}

where $\lambda$ is drawn from a Beta distribution defined by an hyperparameter $\alpha$: $\lambda \sim Beta(\alpha,\alpha)$. This augmentation strategy encourages the trained models to have a linear behavior in-between training samples \cite{mixup},\cite{manifold_mixup}. In practice, Mixup reduces the generalization error of classifiers, makes neural networks less sensitive to corrupted labels and slightly improves the robustness to adversarial examples. In the ablation study carried out in Section \ref{ablation_study}, we detail the influence of Mixup on neural network robustness.

Augmenting datasets by interpolating samples of training sets have been largely studied \cite{between_samples},\cite{sample_pairing}. Among the most successful, the Manifold Mixup interpolates hidden representations instead of interpolating the inputs directly \cite{manifold_mixup}. The Mixup Inference proposes to use Mixup in the inference phase to degrade the perturbations that may corrupt the input images \cite{mixup_inference}. Directional Adversarial Training and Untied Mixup are alternative policies to pick the interpolation ratios of the \textit{mixuped} samples and labels \cite{DAT}. The proposed M-TLAT algorithm uses the standard Mixup, but it is not incompatible with the other interpolation strategies mentioned in this paragraph.

\subsection{Targeted Labeling Adversarial Training: TLAT}

M-TLAT relies on a second data augmentation procedure: adversarial training, which consists in adding adversarial examples into the training set \cite{harnessing},\cite{pgd_aug}. It is one of the most efficient defenses against adversarial examples \cite{obfu_gradient}. We consider an unperturbed sample $x_{clean}$ of size $S$ and its label $y_{clean}$. We can corrupt $x_{clean}$ to build an adversarial example $x_{adv}$:

\begin{equation}\label{untarget_adv_training}
\begin{split}
x_{adv} & = x_{clean} + \underset{\delta \in [-\epsilon,\epsilon]^S}{arg\,max}\{L(x_{clean}+\delta,y_{clean},\theta)\} \\
y_{adv} & = y_{clean}
\end{split}
\end{equation}

Where $\theta$ are the parameters of the attacked model. $L$ is a cost function like the cross-entropy function. The value $\epsilon$ defines the amount of the introduced adversarial perturbation. As suggested in \cite{adv_phy}, adversarial training is even more efficient when it uses adversarial examples that target a specific class $y_{target}$. The augmentation strategy becomes:

\begin{equation}\label{targeted_adv_training}
\begin{split}
x_{adv} & = x_{clean} - \underset{\delta \in [-\epsilon,\epsilon]^S}{arg\,max}\{L(x_{clean}+\delta,y_{target},\theta)\} \\
y_{adv} & = y_{clean}
\end{split}
\end{equation}

One advantage to use target adversarial examples during training is to prevent label leaking \cite{adv_scale}. We propose to improve this augmentation strategy by using $y_{target}$ in the labeling of the adversarial examples. In particular, we propose to mix the one-hot encoding ground-truth labels of the original samples with the one-hot encoding target labels used to craft the adversarial examples:

\begin{equation}\label{targeted_labeling_adv_training}
\begin{split}
x_{adv} & = x_{clean} - \underset{\delta \in [-\epsilon,\epsilon]^S}{arg\,max}\{L(x_{clean}+\delta,y_{target},\theta)\} \\
y_{adv} & = (1-\epsilon)*y_{clean} + \epsilon*y_{target}
\end{split}
\end{equation}

We call this augmentation strategy Targeted Labeling Adversarial Training (TLAT). The $arg\,max$ part is approximated with an adversarial example algorithm such as target FGSM \cite{adv_phy}:

\begin{equation} \label{target_fgsm}
x_{adv} = x_{clean} - \epsilon*sign(\nabla_{x_{clean}}L(x_{clean},y_{target},\theta))
\end{equation}

FGSM is used because it is a computationally efficient way to craft adversarial examples \cite{harnessing},\cite{ensemble_training}. As for traditional adversarial trainings, models trained with TLAT should be trained on both clean samples and adversarial samples \cite{adv_phy}. The advantage of TLAT is to make models have a high clean accuracy compared to the models trained with a standard adversarial training algorithm. More details can be found in Section \ref{adv_study}.\\

TLAT uses soft labels instead of one-hot encoding labels. Using soft labels is a recurrent idea in the methods that address robustness. As mentioned above, Mixup interpolates training labels to generate soft labels \cite{mixup}. Label smoothing replaces the zeros of one-hot encoding labels by a smoothing parameter $s > 0$ and normalizes the high value so that the distribution still sums to one \cite{label_smoothing}. Distillation learning uses the logits of a trained neural network to train a second neural network \cite{distillation}. The second network is enforced to make smooth predictions by learning on soft labels. Bilateral training generates soft labels by adversarially perturb training labels \cite{bilateral}. It uses the gradient of the cost function of an attacked model to generate adversarial labels. Models trained on both adversarial examples and adversarial labels are encouraged to have a small gradient magnitude which makes them more robust to adversarial examples. 

The originality of TLAT is to use the target labels of adversarial attacks as a component of the soft labels. Intuitions about why this labeling strategy works are provided in Section \ref{interpretation TLAT}.

\subsection{M-TLAT}

The idea of M-TLAT is to generate new training couples by applying sequentially the TLAT perturbations (\ref{targeted_labeling_adv_training}) after the Mixup interpolations (\ref{mixup}): 

\begin{equation}\label{ep:mtlat}
\begin{split}
x_{mtlat} & = x_{mix} - \underset{\delta \in [-\epsilon,\epsilon]^S}{arg\,max}\{L(x_{mix}+\delta,y_{target},\theta)\} \\
y_{mtlat} & = (1-\epsilon)*y_{mix} + \epsilon*y_{target}
\end{split}
\end{equation}

As displayed in Figure \ref{fig}, $x_{mtlat}$ contains features that come from three distinct sources: two clean images and an adversarial perturbation that targets a specific class. The label $y_{mtlat}$ contains the class and the weight associated with each source of features. These weights are determined by the values $\lambda$ and $\epsilon$. A model trained with M-TLAT is not only constraint to predict the classes that correspond to the three sources. It also has to predict the weight of each source within the features of $x_{mtlat}$. We believe that being able to predict the class and the weighting of the three sources requires a subtle understanding of the features present in images. In practice, being trained with augmented couples $(x_{mtlat},y_{mtlat})$ makes neural networks more robust.\\

The expressions (\ref{ep:mtlat}) are the essence of the algorithm. The whole process of one training step with M-TLAT is provided in the algorithm description \ref{alg:algorithm}. As recommended in \cite{adv_phy}, the training minibatches contain both adversarial and non-adversarial samples. In our algorithm, the non-adversarial samples are obtained using a standard Mixup interpolation, the adversarial samples are crafted by combining Mixup and TLAT.

Another recently proposed approach combines Mixup and adversarial training \cite{interpol_adv}. Their method called Interpolated Adversarial Training (IAT) is different from M-TLAT for two main reasons. Most importantly, they do not use the labeling strategy of TLAT. Basically, their adversarially corrupted samples are labeled using the standard Mixup interpolation while our labels contain information about the amount and the target of the used adversarial examples. Secondly, we interpolate images before adding the adversarial corruptions. On the contrary they adversarially corrupt images before mixing them up. In practice, we get better adversarial robustness when we proceed in our order. Besides, proceeding in their order doubles the number of adversarial perturbations to compute: it increases the training time. In Section \ref{iat_vs_M-TLAT}, we compare the robustness of two models trained with these approaches.

\begin{algorithm} 
\caption{One training step of the M-TLAT algorithm}
\label{alg:algorithm}
\begin{algorithmic}
\REQUIRE $\theta$ the parameters of the trained neural network
\REQUIRE $L$ a cross entropy function
\REQUIRE $(x_{1},y_{1}),(x_{2},y_{2}),(x_{3},y_{3}),(x_{4},y_{4}) \sim$ Dataset
\REQUIRE $\lambda_{1},\lambda_{2} \sim Beta(\alpha,\alpha)$
\REQUIRE $\epsilon \sim U[0,\epsilon_{max}]$ where $U$ is a uniform distribution and $\epsilon_{max}$ is the maximum perturbation allowed
\REQUIRE $y_{target} \sim U[0,N]$ where  N is the number of classes
\REQUIRE $optim$ an optimizer like Adam or SGD

\STATE $x_{mix1} = \lambda_{1}*x_{1} + (1-\lambda_{1})*x_{2}$
\STATE $y_{mix1} = \lambda_{1}*y_{1} + (1-\lambda_{1})*y_{2}$
\STATE
\STATE $x_{mix2} = \lambda_{2}*x_{3} + (1-\lambda_{2})*x_{4}$
\STATE $y_{mix2} = \lambda_{2}*y_{3} + (1-\lambda_{2})*y_{4}$
\STATE
\STATE $x_{mtlat} = x_{mix2} - \epsilon*sign(\nabla_{x_{mix2}}L(x_{mix2},y_{target},\theta))$
\STATE $y_{mtlat} = (1-\epsilon)*y_{mix2} + \epsilon*y_{target}$
\STATE
\STATE $ loss_{1} = L(x_{mix1},y_{mix1},\theta)$
\STATE $ loss_{2} = L(x_{mtlat},y_{mtlat},\theta)$
\STATE $loss = loss_{1} + loss_{2}$
\STATE $gradients = \nabla_{\theta}loss$
\STATE $optim.update(gradient, \theta)$ The optimizer updates $\theta$ according to the gradients
\end{algorithmic}
\end{algorithm}

\section{Experiment Set-up}

\subsection{Perturbation Benchmark} \label{benchmark}

We want to evaluate the robustness of neural networks in a broad sense, covering the most encountered common corruptions and adversarial examples. To achieve this, we gather a large set of perturbations that contains nineteen common corruptions and five adversarial attacks.

\subsubsection{Common Corruptions.} The benchmark includes the common corruptions of ImageNet-C \cite{imagenet_c}. The ImageNet-C set of perturbations contains diverse kinds of common corruptions such as 1) noises: Gaussian noise, shot noise and impulse noise 2) blurs: defocus blur, glass blur, motion blur and zoom blur 3) weather related corruptions: snow, frost, fog and brightness 4) digital distortions: contrast, elastic, pixelate and jpeg compression. Each corruption is associated with five severity levels. We use the corruption functions\footnote{https://github.com/hendrycks/robustness} provided by the authors to generate those perturbations.
 
The rotation, translation, color distortion and occlusion common corruptions are absent from ImageNet-C, yet they are often encountered in industrial applications. We decided to add those four perturbations to the pool of common corruptions. The occlusion perturbation is modeled by masking a randomly selected region of images with a grey square. For the images corrupted with color distortion, one of the RGB channel is randomly chosen and a constant is added to all the pixels of this channel. For the rotation and translation perturbations, the pixel values of the area outside the transformed images are set to zero.

These four corruptions are associated with a severity range. The lower bound of the severity range has been set to reduce the accuracy of the standard ResNet-50 by five percent on the test set. The upper bound has been set to reduce its accuracy by thirty percent. As a result, the rotations can turn images from 8 to 40 degrees. In other words, when the images of the test set are rotated by eight degrees, the accuracy of the standard ResNet-50 is five percent lower than for a not-corrupted test set. The side of the square mask used in the occlusions varies from 60 to 127 pixels. The translations can move images from 15 to 62 pixels. For the color distortion corruption, the values of the pixels of the perturbed channel are increased from 8\% to 30\% of the maximum possible pixel value.

\subsubsection{Adversarial Examples.} 
Following the recommendations in \cite{evaluating}, we carefully choose the adversarial examples to be added to the benchmark. Firstly, we use adversarial examples in both white box and black box settings. Secondly, we use two different metrics, the $L_{\infty}$ and the $L_{2}$ norms, to compute the bound of adversarial attacks. Thirdly, we employ targeted and untargeted attacks. Fourthly, several amounts of adversarial perturbations are used. Finally, the selected adversarial examples are not used during trainings. We build a set of adversarial examples to cover all the conditions mentioned above. 

We note $\epsilon$ the amount of the introduced perturbation in adversarial examples. We use PGD with $\epsilon = 0.04$ as a white-box $L_{\infty}$ bounded attack \cite{pgd_aug}. We generate targeted adversarial attacks by using PGD\_LL, which targets the least likely class according to attacked models \cite{adv_phy}. We use MI\_FGSM with $\epsilon = 0.04$ and $\epsilon = 0.08$ as black-box attacks \cite{MI_FGSM}. A VGG network trained on the same training set is used to craft these black-box attacks. PGD, PGD\_LL and MI\_FGSM are computed over ten iterations. We use the Carlini-Wagner attack ($CW_2$) as a $L_{2}$ white-box bounded attack \cite{c&w}. We perform the optimization process of this attack with 40 iterations and a confidence score of 50.

The gathered common perturbations and adversarial examples constitute the perturbation benchmark. It is used in the experimental section in order to evaluate and compare the robustness of models.

\subsection{Training Details}

The trained models are either a ResNet-50 \cite{resnet} or a ResNeXt-50 with the 32x4d template \cite{resnext}. We use a batch size of 256 and 90 training epochs. The Adam optimizer \cite{adam} is used with a learning rate of 0.002 and a weight decay of $10^{-4}$. At the end of the epochs 30, 60 and 80, the learning rate is divided by 10. The cost function is the cross entropy. We use the same hyperparameters for all trainings.

All models are trained and tested using ImageNet. Because of a limited computational budget, we used a subset of ImageNet built on 100 randomly chosen classes. For each class, ten percent of the images are preserved for the test set. Then, the training set and the test set contain respectively $10^{5}$ and $10^{4}$ images. The only pre-processing used is the resizing of images to the 224*224 format.

We call the models trained without any data augmentation the standard models. We observed that the highest clean accuracy for the models trained with mixup is reached when $\alpha = 0.4$, so we used this value in all experiments. The adversarial examples used in trainings are crafted using FGSM with $\epsilon \sim U[0,0.025]$. The range of pixel values of images in our experiments is $[0,1]$.

\section{Performance Evaluation}

\subsection{Robustness Score}

To measure the robustness of a model to a perturbation, we compare the performance of this model on clean samples with its performance on perturbed samples:

\begin{equation} \label{robust_expr}
R^{\phi}_{N} = \frac{A_{\phi}}{A_{clean}}
\end{equation}

We call $R^{\phi}_{N}$ the robustness score of a neural network $N$ towards a perturbation $\phi$. $A_{clean}$ is the accuracy of the model on the clean test set and $A_{\phi}$ is its accuracy on the test set corrupted with the $\phi$ perturbation. $\phi$ can be either a common corruption or an adversarial attack.

This robustness score metric should be used carefully because it masks the clean accuracy of neural networks. Indeed, an untrained model that always makes random outputs, would have the same accuracy for clean samples than for corrupted samples. Its robustness scores would be equal to 1, so it could be considered as completely robust to any common corruption. Therefore, in this study, before comparing the robustness scores of two neural networks we always make sure that their clean accuracies are also comparable.

\subsection{Performances of M-TLAT on the Perturbation Benchmark} \label{iat_vs_M-TLAT}

For the first experiment we train one Resnet-50 and one ResNeXt-50, using the M-TLAT algorithm. The training of these models took a dozen of hours using a single GPU Nvidia Tesla V100. We also train one Resnet-50 and one ResNeXt-50 with the IAT algorithm \cite{interpol_adv}. We compute the robustness scores of the trained models towards all the perturbations of the benchmark. The results are reported in Table \ref{tab:main_res}. 

In Tables \ref{tab:main_res} and \ref{tab:ablation_study}, the \textit{Clean} column contains the accuracy of the models on the not-corrupted test set. Each of the other columns contains the robustness scores to a perturbation of the benchmark. For the corruptions of the ImageNet-C benchmark, the displayed scores correspond to the mean robustness score of the corruptions computed with their five severity levels. To better visualize the effect of the augmentation algorithms, we use either a "-" index or a "+" index, to signify if a model is less or more robust than the standard model to a perturbation. 

We observe in Table \ref{tab:main_res} that the models trained with M-TLAT are slightly more accurate than the standard models on clean images. They are also more robust than the standard models to every single tested common corruption. We see that using M-TLAT makes neural networks much more robust to the $CW_{2}$ and PGD\_LL attacks. It also makes models less sensitive to black-box adversarial examples. We observe that the robustness gain for the PGD attack is less important.

For comparison, the IAT algorithm tends to reduce the clean accuracy. It does not increase the robustness to all the common corruption of the benchmark. In particular, it significantly decreases the robustness towards the Jpeg perturbation. Besides, the IAT models are significantly less robust to adversarial examples than the M-TLAT models.

Using FGSM during training is known to poorly defend models against iterative adversarial attack such as PGD \cite{adv_scale}. The robustness of the M-TLAT models towards PGD can likely be increased by replacing FGSM by an iterative adversarial attack \cite{pgd_aug}. But this would increase significantly the training time. That is the reason why this option has not been tested yet.

To our knowledge, M-TLAT is the first data augmentation approach that is able to increase the robustness to every single common corruption and adversarial example of a large set of diverse perturbations, without reducing the clean accuracy.

\begin{table}[t]
\caption{Effect on robustness of M-TLAT and comparison with IAT \label{tab:main_res}}
\begin{subtable}{\textwidth}
\caption{Robustness scores towards the common corruptions of ImageNet-C \label{tab:imagenet-c}}
\begin{small}
\resizebox{\textwidth}{!}{%
\begin{tabular}{p{14mm}C{13mm}|C{10mm}|C{10mm}C{10mm}C{10mm}C{10mm}C{10mm}C{10mm}C{10mm}C{10mm}C{10mm}C{10mm}C{10mm}C{10mm}C{10mm}C{10mm}C{10mm}}
\toprule
{} & {} &  Clean &  Gauss &  Shot &  Impul &  Defocus &  Glass &  Motion &  Zoom &  Snow &   Fog &  Frost &  Bright &  Contr &  Elastic &  Pixelate &  Jpeg \\
\midrule
ResNet & standard    &  73.3 &            0.17 &        0.17 &           0.12 &          0.25 &        0.35 &         0.41 &       0.47 &  0.31 &  0.48 &   0.34 &        0.78 &      0.34 &               0.73 &      0.65 &              0.80 \\
& IAT &  $73.2\bm{^{-}}$ &            $0.47^{+}$ &        $0.46^{+}$ &           $0.42^{+}$ &          $0.51^{+}$ &        $0.65^{+}$ &         $0.58^{+}$ &       $0.68^{+}$ &  $0.49^{+}$ &  $0.78^{+}$ &   $0.66^{+}$ &       $0.79^{+}$ &      $0.78^{+}$ &              $0.84^{+}$ &      $0.82^{+}$ &              $0.63\bm{^{-}}$ \\
& M-TLAT  &  $73.9^{+}$ &            $0.56^{+}$ &        $0.54^{+}$ &           $0.52^{+}$ &          $0.41^{+}$ &        $0.61^{+}$ &         $0.58^{+}$ &       $0.63^{+}$ &  $0.49^{+}$ &  $0.59^{+}$ &   $0.61^{+}$ &        $0.82^{+}$ &      $0.58^{+}$ &               $0.85^{+}$ &      $0.94^{+}$ &              $0.95^{+}$ \\\cmidrule{1-18}
ResNeXt & standard &  76.4 &           0.25 &       0.25 &          0.20 &         0.28 &       0.37 &        0.44 &      0.48 &  0.36 &  0.53 &  0.37 &       0.79 &     0.36 &              0.75 &     0.72 &             0.74  \\
& IAT &  $74.7\bm{^{-}}$ &           $0.46^{+}$ &       $0.44^{+}$ &          $0.43^{+}$ &         $0.53^{+}$ &       $0.67^{+}$ &        $0.62^{+}$ &      $0.70^{+}$ &  $0.51^{+}$ &  $0.73^{+}$ &  $0.69^{+}$ &       $0.81^{+}$ &     $0.80^{+}$ &              $0.85^{+}$ &     $0.83^{+}$ &             $0.59\bm{^{-}}$ \\
& M-TLAT   &  $76.5^{+}$ &           $0.57^{+}$ &       $0.55^{+}$ &          $0.54^{+}$ &         $0.44^{+}$ &       $0.64^{+}$ &       $0.60^{+}$ &     $0.66^{+}$ &  $0.52^{+}$ &  $0.68^{+}$ &  $0.67^{+}$ &       $0.86^{+}$ &     $0.70^{+}$ &              $0.85^{+}$ &     $0.95^{+}$ &             $0.95^{+}$ \\
\bottomrule
\end{tabular}}
\end{small}
\end{subtable}

\begin{subtable}{0.44\linewidth}
\caption{Robustness scores towards our additional common corruptions \label{tab:home_pert}}
\begin{tiny}
\begin{tabular}{p{9mm}C{10mm}|C{8mm}C{8mm}C{7mm}C{8mm}C{8mm}}
\toprule
{} &  {} &  Obstru &  Color &  Trans &  Rot\\
\midrule
ResNet & standard   &   0.74 &            0.76 &        0.78 &     0.68\\
& IAT &  $0.71\bm{^{-}}$ &            $0.89^{+}$ &        $0.75\bm{^{-}}$ &     $0.72^{+}$\\
& M-TLAT  &  $0.75^{+}$ &            $0.86^{+}$ &        $0.79^{+}$ &     $0.74^{+}$\\\cmidrule{1-6}
ResNeXt & standard  &  0.75 &            0.82 &        0.82 &     0.72 \\
& IAT &  $0.72\bm{^{-}}$ &     $0.90^{+}$ &        $0.77\bm{^{-}}$ &     $0.71^{-}$  \\
& M-TLAT  &  $0.76^{+}$ &         $0.89^{+}$ &     $0.82$ &    $0.74^{+}$  \\
\bottomrule
\end{tabular}
\end{tiny}
\end{subtable}%
\hfill
\begin{subtable}{0.53\linewidth}
\caption{Robustness scores towards adversarial examples \label{tab:adv}}
\begin{tiny}
\begin{tabular}{p{9mm}C{10mm}|C{8mm}C{8mm}C{7mm}C{8mm}C{8mm}}
\toprule
{} & {} & pgd $\epsilon$=0.04 &  pgd\_ll $\epsilon$=0.04 & cw\_l2 &  mi\_fgsm $\epsilon$=0.04 &  mi\_fgsm $\epsilon$=0.08 \\ 
\midrule
ResNet & standard    &  0.00 &   0.00 &  0.00 &    0.58 &     0.34 \\
& IAT &  $0.01^{+}$ &   $0.08^{+}$ &  $0.84^{+}$ &    $0.87^{+}$ &     $0.78^{+}$ \\
& M-TLAT  & $0.08^{+}$ &   $0.45^{+}$ &  $1.00^{+}$ &    $0.96^{+}$ &     $0.87^{+}$ \\\cmidrule{1-7}
ResNeXt & standard    &  0.00 &   0.00 &  0.00 &    0.58 &     0.33 \\
& IAT &  $0.01^{+}$ &   $0.11^{+}$ &  $0.95^{+}$ &    $0.87^{+}$ &     $0.81^{+}$ \\
& M-TLAT  &  $0.09^{+}$ &   $0.38^{+}$ &  $0.99^{+}$ &    $0.95^{+}$ &     $0.87^{+}$ \\
\bottomrule
\end{tabular}
\end{tiny}
\end{subtable}
\end{table}

\subsection{Complementarity between Mixup and TLAT}\label{ablation_study}

To better understand the effect of each constituent of the M-TLAT algorithm, we proceed to an ablation study. Two ResNet-50 are respectively trained with the Mixup and TLAT data augmentations. We report their robustness to the perturbation benchmark in Table \ref{tab:ablation_study}.

First, we notice that Mixup causes an increase of the clean accuracy, which is coherent with observations made in \cite{mixup}. On the contrary, TLAT makes the trained model less accurate on the clean data. But those two effects seem to cancel each other because the M-TLAT model and the standard model have comparable clean accuracies as observed in Table \ref{tab:imagenet-c}.

In Tables \ref{tab:imagenet-c_ablation} and \ref{tab:home_pert_ablation}, we observe that Mixup makes the trained model more robust than the standard model to all the common corruptions but the \textit{Motion Blur}, \textit{Pixelate} and \textit{Jpeg} corruptions. We observe in Table \ref{tab:adv_ablation} that Mixup has a little influence on adversarial robustness, with either a slight increase or decrease of the robustness depending on the considered attack.

Fortunately, TLAT makes models much more robust to any adversarial attacks. Indeed, the TLAT model is much more difficult to attack with the black box adversarial examples or with the $CW_2$ attack. It is also significantly more robust to PGD\_LL and slightly more robust to PGD. For common corruptions, the effect of TLAT is very contrasted. Concerning the noise and blur corruptions (the seven first corruptions of the Table \ref{tab:imagenet-c_ablation}), the TLAT model is much more robust than the standard model. For some other common corruptions like \textit{Fog} or \textit{Contrast}, the TLAT augmentation decreases significantly the robustness scores.

It is clear that the M-TLAT models are much more robust to adversarial examples thanks to the contribution of TLAT. However, for common corruptions, Mixup and TLAT are remarkably complementary. Concerning the few corruptions for which Mixup has a very negative effect on robustness (\textit{Jpeg} and \textit{Pixelate}), TLAT has a strong positive effect. Similarly, for the \textit{Fog} and \textit{Contrast} corruptions, TLAT makes models less robust while Mixup makes them much more robust.

The ablation study indicates that both components are important to increase the robustness to a large diversity of perturbations.

\begin{table}[t]
\caption{Influence on robustness of the Mixup and TLAT data augmentations \label{tab:ablation_study}}
\begin{subtable}{\textwidth}
\caption{Robustness scores towards the corruptions of ImageNet-C \label{tab:imagenet-c_ablation}}
\begin{small}
\resizebox{\textwidth}{!}{%
\begin{tabular}{p{14mm}C{13mm}|C{10mm}|C{10mm}C{10mm}C{10mm}C{10mm}C{10mm}C{10mm}C{10mm}C{10mm}C{10mm}C{10mm}C{10mm}C{10mm}C{10mm}C{10mm}C{10mm}}
\toprule
{} & {} &  Clean &  Gauss &  Shot &  Impul &  Defocus &  Glass &  Motion &  Zoom &  Snow &   Fog &  Frost &  Bright &  Contr &  Elastic &  Pixelate &  Jpeg \\
\midrule
ResNet & standard    &  73.3 &            0.17 &        0.17 &           0.12 &          0.25 &        0.35 &         0.41 &       0.47 &  0.31 &  0.48 &   0.34 &        0.78 &      0.34 &               0.73 &      0.65 &              0.80  \\
& Mixup  &  $74.9^{+}$ &            $0.28^{+}$ &        $0.28^{+}$ &           $0.24^{+}$ &          $0.25^{ }$ &        $0.38^{+}$ &         $0.39^{-}$ &       $0.53^{+}$ &  $0.38^{+}$ &  $0.79^{+}$ &   $0.53^{+}$ &        $0.78^{ }$ &      $0.75^{+}$ &               $0.75^{+}$ &      $0.58^{-}$ &              $0.61^{-}$ \\
& TLAT &  $69.4^{-}$ &            $0.57^{+}$ &        $0.54^{+}$ &           $0.51^{+}$ &          $0.43^{+}$ &        $0.60^{+}$ &         $0.56^{+}$ &       $0.60^{+}$ &  $0.41^{+}$ &  $0.15^{-}$ &   $0.41^{+}$ &        $0.78^{ }$ &      $0.13^{-}$ &               $0.84^{+}$ &      $0.94^{+}$ &              $0.97^{+}$  \\
\bottomrule
\end{tabular}}
\end{small}
\end{subtable}

\begin{subtable}{0.44\linewidth}
\caption{Robustness scores towards our additional common corruptions \label{tab:home_pert_ablation}}
\begin{tiny}
\begin{tabular}{p{8mm}C{9mm}|C{8mm}C{8mm}C{8mm}C{8mm}C{8mm}}
\toprule
{} &  {} &  Obstru &  Color &  Trans &  Rot\\
\midrule
ResNet & standard  &   0.74 &            0.76 &        0.78 &     0.68\\
& Mixup   &   $0.75^{+}$ &            $0.88^{+}$ &        $0.79^{+}$ &    $0.71^{+}$  \\
& TLAT  &   $0.69^{-}$ &            $0.76^{ }$ &        $0.67^{-}$ &     $0.66^{-}$ \\
\bottomrule
\end{tabular}
\end{tiny}
\end{subtable}%
\hfill
\begin{subtable}{0.53\linewidth}
\caption{Robustness scores towards adversarial examples \label{tab:adv_ablation}}
\begin{tiny}
\begin{tabular}{p{8mm}C{9mm}|C{8mm}C{8mm}C{8mm}C{8mm}C{8mm}}
\toprule
{} & {} & pgd $\epsilon$=0.04 &  pgd\_ll $\epsilon$=0.04 & cw\_l2 &  mi\_fgsm $\epsilon$=0.04 &  mi\_fgsm $\epsilon$=0.08 \\ 
\midrule
ResNet & standard    &  0.00 &   0.00 &  0.00 &    0.58 &     0.34 \\
& Mixup &  0.00 &   0.00 &  $0.202^{+}$ &    $0.61^{+}$ &     $0.22^{-}$ \\
& TLAT &  $0.10^{+}$ &  $0.74^{+}$ & $0.97^{+}$ &    $0.98^{+}$ &     $0.93^{+}$ \\
\bottomrule
\end{tabular}
\end{tiny}
\end{subtable}
\end{table}

\subsection{Labeling in Adversarial Trainings} \label{adv_study}

\subsection*{Comparison of the Labeling Strategies}

Adversarial trainings increase the adversarial robustness of the trained models, but they also reduce their accuracy on clean samples \cite{odds_with},\cite{adv_scale}. In this section, we want to show that TLAT decreases less the clean accuracy of the trained models than traditional adversarial trainings.

To achieve it, we trained four ResNet-50 with different kinds of adversarial training algorithms. The first model is trained using untarget FGSM and the second is trained using target FGSM with randomly chosen target. Both use adversarial examples labeled with the ground-truth labels. We train another model with target FGSM but regularized via label smoothing (LS) \cite{label_smoothing}, we call it the LS model. For this model, we use a smoothing parameter equal to $\epsilon$, where $\epsilon$ is the amount of the FGSM perturbation. In other words, the one values of the one-hot encoding vectors are replaced by $1-\epsilon$ and the zeros are replaced by $\epsilon/N$ where N is the number of classes. The fourth model is trained using the TLAT algorithm. All models are trained with minibatches that contain both clean samples and adversarial examples. We measure the clean accuracy of those models: results are displayed in Table \ref{tab:adv_study}. 

We see that TLAT is the adversarial training method that reduces the less the clean accuracy. This result shows that TLAT is important to preserve the clean accuracy of the models trained with M-TLAT, all the while making them more robust to adversarial examples. 

Using soft labels in trainings is known to help models to generalize \cite{distillation},\cite{label_smoothing}. Here we want to make sure that the usage of soft labels is not the main reason of high clean accuracy of TLAT. To achieve it, we compare the performances of the TLAT and LS models. Even if the LS model also uses soft labels during training, it performs worse than the TLAT model. Consequently, the good performances of TLAT are not due to the usage of soft labels. We believe TLAT performs well because it uses labels that contain information about the target of the adversarial examples.

\begin{table}
\begin{center}
\caption{Comparison of the performances of the TLAT augmentation with the performances of other kinds of adversarial trainings \label{tab:adv_study}}
\begin{tabular}{p{22mm}|C{18mm}|C{18mm}C{20mm}C{18mm}C{18mm}}
\toprule
{} & standard & FGSM & target-FGSM & LS  & TLAT \\\cmidrule{1-6}
clean accuracy & 73.3 & 65.8 & 68.3 & 67.1 & 69.4 \\
\bottomrule
\end{tabular}
\end{center}
\end{table}

\subsection*{Interpretation} \label{interpretation TLAT}

The TLAT augmentation is motivated by the works of Ilyas et al \cite{adv_features}. In their study, they reveal the existence of brittle yet highly predictive features in the data. Neural networks largely depend on those features even if they are nearly invisible to human eye. They are called non-robust features. Ilyas et al. show that a model that only uses non-robust features to complete properly a task still generalize well on unseen data. 

They use this phenomenon to interpret adversarial vulnerability. Adversarial attacks are small perturbations, so they mainly affect non-robust features. They can especially make those brittle features anti-correlated with the true label. As neural networks largely rely on non-robust features, their behaviour is completely disturbed by adversarial attacks.

In Adversarial trainings, neural networks encounter adversarial examples that can have non-robust features uncorrelated with the true label. The trained models are then constrained to less rely on non-robust features. This can explain the success of adversarial training: adversarially trained models give less importance to non-robust features so they are much more difficult to attack with small perturbations.

Despite its efficiency, adversarial training generally causes a decrease in clean accuracy \cite{odds_with},\cite{adv_scale},\cite{trades}. One possible reason could be that adversarial patterns in the training adversarial examples are not coherent with the ground-truth label. Consequently, an adversarially trained model encounters samples for which the features are not completely coherent with labelling.

The proposed method tries to make the labeling of target adversarial examples more correlated with its non-robust features. Targeted adversarial attacks make non-robust features of a sample correlated with a target class. So instead of labelling only with the ground-truth class of the attacked sample, the method introduces a part related to the target class. Therefore, the trained model still learns the true original class of the attacked sample, but the label used for learning is more correlated with the non-robust features of this sample. We believe this could be the reason why our method has better performance than traditional target adversarial training in practice.

\section{Conclusion}

We propose a new data augmentation strategy that increases the robustness of neural networks to a large set of common corruptions and adversarial examples. The experiments carried out suggest that the effect of M-TLAT is always positive: basically, it increases the robustness to any corruption without reducing the clean accuracy. We believe using M-TLAT can be particularly useful to help industrials to increase the robustness of their neural networks without being afraid of any counterpart.

As part of the M-TLAT algorithm, we use the new adversarial augmentation strategy TLAT. We show that models trained with TLAT have a better accuracy on clean samples than the models trained with a standard adversarial training algorithm. The idea of TLAT is to interpolate the target labels of adversarial examples with the ground-truth labels. This operation is computationally negligible and can be used in any trainings with target adversarial examples in order to improve the clean accuracy.

In future works, we would like to replace FGSM by an iterative adversarial attack in our algorithm and observe how this would influence the clean accuracy and the adversarial robustness of the models trained with M-TLAT. It would be also interesting to replace Mixup by a different interpolation strategy such as Manifold Mixup \cite{manifold_mixup}, and test if it further improves the performances of the algorithm.

\bibliographystyle{splncs04}
\bibliography{egbib}
\end{document}